\documentclass[10pt,twocolumn,letterpaper]{article}

\usepackage{cvpr}              
\definecolor{cvprblue}{rgb}{0.21,0.49,0.74}
\usepackage[pagebackref,breaklinks,colorlinks,allcolors=cvprblue]{hyperref}
\usepackage{booktabs}
\usepackage{tabularx}
\usepackage{threeparttable}
\usepackage{multirow}
\usepackage{graphicx}
\usepackage{subcaption}
\usepackage{tcolorbox}
\tcbuselibrary{skins, breakable}
\usepackage{algorithm}
\usepackage{algorithmic}

\usepackage{amsmath}
\usepackage{amssymb}
\usepackage{diagbox} 
\def\confName{CVPR}
\def\confYear{2026}

\title{Iterative Closed-Loop Motion Synthesis for Scaling the Capabilities of Humanoid Control}

\author{
    Weisheng Xu$^{1*}$, \quad 
    Qiwei Wu$^{1*}$, \quad 
    Jiaxi Zhang$^1$, \quad 
    Tan Jing$^1$, \quad 
    Yangfan Li$^1$, \\
    Yuetong Fang$^1$, \quad 
    Jiaqi Xiong$^2$, \quad 
    Kai Wu$^1$, \quad 
    Rong Ou$^1$, \quad 
    Renjing Xu$^{1\dagger}$ \\
    \\
    $^1$Hong Kong University of Science and Technology (Guangzhou) \\
    $^2$University of Oxford \\
    {\tt\small \{wxu421, qwu720\}@connect.hkust-gz.edu.cn, renjingxu@hkust-gz.edu.cn}
}

\begin{document}
\maketitle

\def\thefootnote{*}\footnotetext{Equal contribution.}\def\thefootnote{\arabic{footnote}}
\def\thefootnote{$\dagger$}\footnotetext{Corresponding author.}\def\thefootnote{\arabic{footnote}}
\begin{abstract}
Physics-based humanoid control relies on training with motion datasets that have diverse data distributions. However, the fixed difficulty distribution of datasets limits the performance ceiling of the trained control policies. Additionally, the method of acquiring high-quality data through professional motion capture systems is constrained by costs, making it difficult to achieve large-scale scalability. To address these issues, we propose a closed-loop automated motion data generation and iterative framework. It can generate high-quality motion data with rich action semantics, including martial arts, dance, combat, sports, gymnastics, and more. Furthermore, our framework enables difficulty iteration of policies and data through physical metrics and objective evaluations, allowing the trained tracker to break through its original difficulty limits. On the PHC single-primitive tracker, using only approximately 1/10 of the AMASS dataset size, the average failure rate on the test set (2201 clips) is reduced by 45\% compared to the baseline. Finally, we conduct comprehensive ablation and comparative experiments to highlight the rationality and advantages of our framework.
\end{abstract}

\section{Introduction}
\label{sec:intro}

Physics-based humanoid control supports high dynamic locomotion, character animation, and interactive VR. The standard pipeline imitates MoCap with RL. DeepMimic~\cite{peng2018deepmimic}  showed complex skills are learnable; AMP~\cite{peng2021amp}  improved realism; ASE~\cite{peng2022ase}, DReCon~\cite{bergamin2019drecon}, and PHC~\cite{luo2023perpetual}  advanced reuse and generalization. Progress remains limited by costly expert MoCap and datasets that skew toward low difficulty or narrow domains.

Existing corpora reflect this trade-off. AMASS covers general low dynamic motion, while AIST++ targets professional dance~\cite{li2021ai}. Humanoid-X~\cite{mao2024learning} and HuBE~\cite{lyu2025hube} scale via video mining or cross-embodiment aggregation, but still favor low difficulty or limited scope. Controllers trained on fixed low dynamic distributions often fail on acrobatic motions due to missing high dynamic features. The scarcity of professional, high dynamic data further constrains progress.

\begin{figure*}[h!]
    \centering
    \includegraphics[width=0.95\linewidth]{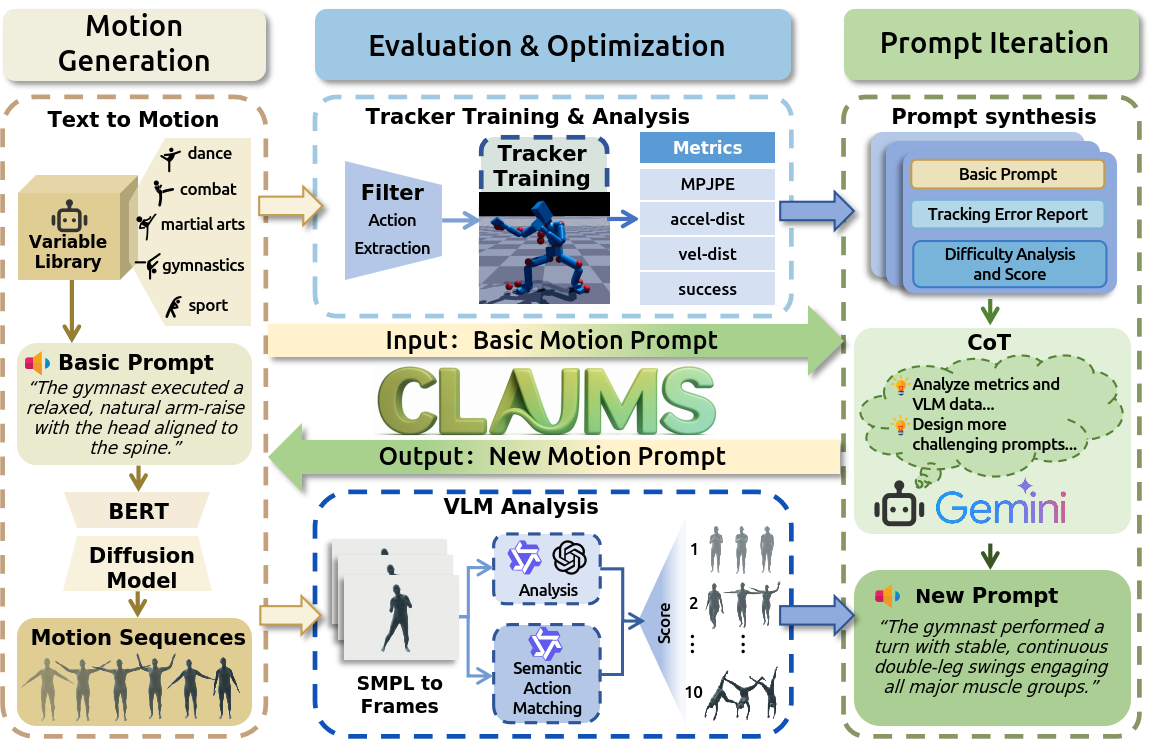}
    \caption{Overview of the CLAIMS pipeline: a closed-loop system that refines prompts from a 5-domain library (martial arts, dance, combat, sports, gymnastics), synthesizes motions with MDM, filters them by physics and VLM checks, trains humanoid trackers with RL, and uses multimodal feedback to generate progressively harder tasks.}
    \label{fig:main}
\end{figure*}

We introduce CLAIMS (as shown in Fig. \ref{fig:main}), a closed-loop automated framework that co-evolves motion data synthesis and controllers. Each iteration generates harder professional motions with motion diffusion model (MDM) from language prompts, trains the controller, analyzes failures with a multimodal agent, and refines data to expand capability.

Our contributions are: (1) A semantic definition and normalization for action data, yielding a scalable dataset with difficulty tiers across diverse professional skills. (2) A competitive iterative procedure that alternates data generation and controller optimization in a game-like setup, enabling the policy to surpass its difficulty ceiling and master high difficulty motions.

After iterative optimization, the dataset exhibits clear difficulty stratification and professional characteristics while preserving motion–text alignment. Only the controller requires training compute; other components are training-free and low cost, making the framework controller agnostic. As shown in Table~\ref{tab:succ vs baseline}, with fewer than 400 training sequences, our tracker surpasses baselines on AIST++, Motion-X/Kungfu~\cite{zhang2025motion}, EMDB~\cite{kaufmann2023emdb}, and Video-Convert~\cite{jing2025farm} on average. Performance improves with more iterations, and the framework consistently boosts diverse trackers. 


\section{RELATED WORK}
\label{sec:formatting}

\paragraph{Humanoid Motion Datasets.}
AMASS~\cite{AMASS} unifies motion under SMPL~\cite{SMPL:2015} but over 90\% of its sequences are low dynamic daily activities, and controllers trained on it generalize poorly to expert actions, as shown by PHC~\cite{luo2023perpetual}. Domain-specific sets such as AIST++~\cite{li2021ai} and EMDB~\cite{kaufmann2023emdb} raise dynamism or annotation fidelity but remain narrow in coverage, scale, and diversity. PHUMA~\cite{lee2025phuma} scales video-based data with physics-constrained retargeting to improve plausibility, yet focuses on locomotion. To reduce collection cost at scale, Humanoid-X~\cite{mao2024learning} and HuBE~\cite{lyu2025hube} mine internet and cross-embodiment motion, but lack reliable semantics and difficulty stratification. Recent million-scale corpora aim to couple scale with semantics and evaluation, for example Go to Zero~\cite{fan2025go} with automated high precision labels and a zero-shot benchmark, MotionLib~\cite{wang2024scaling} with hierarchical text and Motionbook encoding for semantic alignment, and Motion-X++~\cite{zhang2025motion} with expanded multimodal full-body coverage. Despite these advances, most datasets assume a static difficulty distribution and do not provide controller-aware grading or a dynamic expansion process that co-evolves with policy capability. Models for text or multimodal motion generation. MDM~\cite{MDM} established a text-conditioned diffusion baseline; MoFusion~\cite{dabral2023mofusion} improves denoising and sample quality; MotionGPT and MotionGPT-2~\cite{jiang2023motiongpt,wang2024motiongpt} unify generation and understanding with a language model formulation; T2M-X~\cite{liu2024t2m} extends to partially annotated full body motion. For real-time controllability and deployment, Being-M0.5~\cite{cao2025being} leverages the HuMo100M scale corpus and motion tokenization to enable unified vision, language, and motion control. However, these models largely inherit static, low dynamic training distributions and insufficiently cover professional actions; even with scaling and zero-shot evaluation~\cite{fan2025go,wang2024scaling}, they lack difficulty adaptive mechanisms tied to controller mastery.

Contrast to our work. We build a scalable dataset across five professional domains—martial arts, dance, combat, sports, and gymnastics—with semantic tags and explicit difficulty tiers. Our closed-loop framework expands the high difficulty distribution in response to the controller’s current competence, overcoming fixed difficulty corpora and enabling stable gains in professional highly dynamic scenarios.

\paragraph{Physics-Based Humanoid Control.}
Physics-based humanoid control has advanced in tracking yet still faces two bottlenecks: reliance on fixed training distributions and weak generalization across trackers. Early imitation and tracking methods such as DeepMimic and Skeleton2Humanoid~\cite{peng2018deepmimic,li2022skeleton2humanoid} improved skill coverage and plausibility. Adversarial and skill reuse frameworks including AMP, ASE, and PHC~\cite{peng2021amp,peng2022ase,luo2023perpetual} further enhanced realism and coverage, and Neural Categorical Priors~\cite{zhu2023neural} reduced behavioral imbalance. Despite this, controllers inherit static data and struggle on unseen high dynamic skills; universal controllers such as UHC and MaskedMimic~\cite{luo2022universal,tessler2024maskedmimic} often fail on gymnastics tumbling. FARM~\cite{jing2025farm} targets explosive motions with frame-level augmentation and a residual MoE and introduces the HDHM benchmark, yet a gap in high dynamic capability and evaluation remains.

To broaden capability, diffusion and RL have been integrated in a closed loop for multi-task control~\cite{liao2025beyondmimic,truong2024pdp,tevet2024closd}, and language or multimodality has been used for richer task specification~\cite{juravsky2022padl,ding2025humanoid}. These approaches still rely on static corpora and bespoke pipelines and lack difficulty adaptation to controller mastery. At a broader level, zero-shot whole body control~\cite{tirinzoni2025zero} leverages behavioral foundation models but depends on large offline pretraining and does not deliver universally adaptive training for diverse single primitive trackers.
We propose a general training framework for single primitive trackers that couples controller optimization with adaptive difficulty escalation. It raises data difficulty in response to controller competence, requires no specialized high dynamic datasets or pipeline changes, and yields consistent gains on professional high difficulty motion benchmarks.

\paragraph{Iterative Data Generation.} 
Data scarcity remains the main bottleneck across domains, with pipelines still limited in domain specificity, scale, and physical realism. A growing trend is adaptive closed loop curation that couples data selection and generation so models and datasets improve together.

In reasoning and language modeling, SAI-DPO and SEAL~\cite{rao2025dynamic,zweiger2025self} adapt data to the model during training by selecting hard examples, self-generating fine-tuning data, and self-editing instructions. In robotics, RoboTwin 2.0~\cite{chen2025robotwin} uses a multimodal language model pipeline to auto generate tasks and objects and improves sim-to-real transfer, but its loop relies mainly on success rates and image metrics and leaves evaluation under specified. In character control, PARC~\cite{Xu_2025} pairs a motion generator with a physics-based tracker in an iterative generate-correct-augment-retrain loop, yet adopts a single evaluation criterion and does not generalize well across scenarios.

We address these gaps with two components: (1) a professional semantic taxonomy for motion that enables principled categorization and difficulty stratification; and (2) a fused evaluation and selection loop that combines controller performance metrics with vision language model feedback under a multi-dimensional protocol. Guided by the taxonomy, the loop escalates difficulty adaptively, pushes generated motions away from pretraining biases, and provides a clear basis for sustained improvement.

\section{Methodology}
\label{sec:method}

\subsection{Defining High-Difficulty Motion Data}
\label{subsec:pipeline-a}
\begin{figure}
    \centering
    \includegraphics[width=0.85\linewidth]{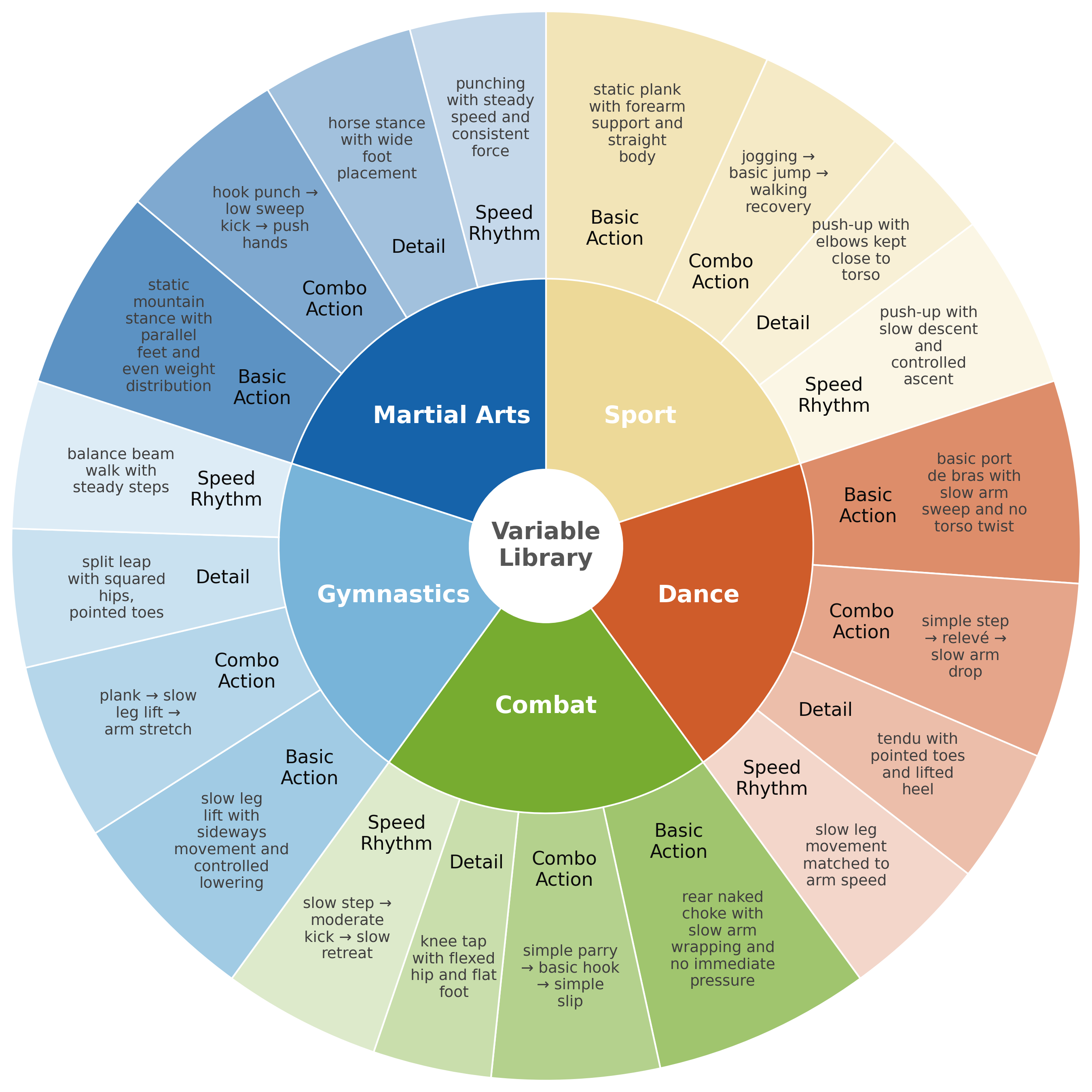}
    \caption{Difficulty-aware variable library across five domains and four compositional axes.}
    \label{fig:varlib}
\end{figure}
\label{subsubsec:gen}



We formalize motion professionalism and difficulty for high-skill controller training. Guided by expert requirements and capture risk, we curate five domains (Fig.~\ref{fig:varlib}): sports, dance, combat, gymnastics, and martial arts, emphasizing high-dynamic motions. Difficulty is defined along four axes: base action (atomic skill); combo action (composition logic); detail (technical cues like limb placement); and speed and rhythm (temporal structure). For example, a dance prompt combines a grand allegro (base), a saut de basque chain (combo), and a triple pirouette (detail) at a steady tempo (speed). These templates constrain generation and guide dataset optimization, ensuring principled difficulty escalation aligned with professional characteristics.

\subsection{Dataset Generation}
\label{subsubsec:track}

We synthesize training data with a low cost, pretrained text conditioned motion diffusion model, MDM~\cite{MDM}, using a 50 step sampler and a DistilBERT text encoder. Although MDM is trained on HumanML3D, its latent space enables compositional mixing of action primitives, yielding behaviors that remain distributionally consistent with the source corpus while introducing novel combinations absent from the original data. Sampling follows standard conditional diffusion with fixed weights. To elicit complex professional behaviors without modifying the generator, we use templated action prompts derived from our semantic taxonomy and automatically instantiate them with an auxiliary LLM. The template-based design provides more stable, domain faithful control over motion composition and difficulty than free form LLM descriptions.

\begin{figure}
    \centering
    \includegraphics[width=0.95\linewidth]{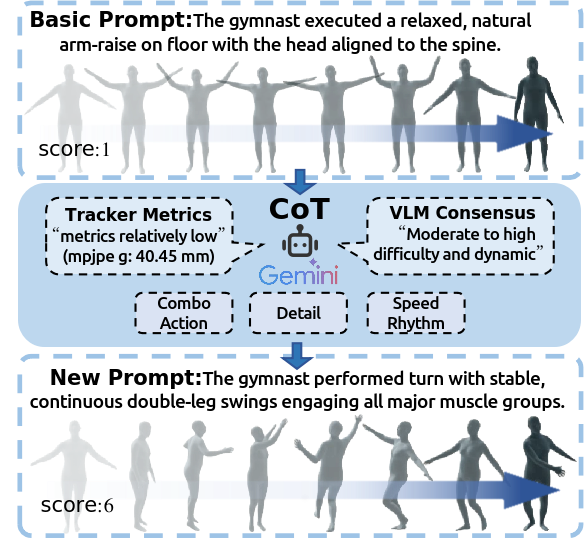}
    \caption{Prompt-to-prompt data generation.}
    \label{fig:p2p}
\end{figure}

In each generation round, we sample text prompts from a difficulty-aware variable bank that spans five expert domains: martial arts, dance, combat, gymnastics, and sports. DistilBERT encodes each prompt into a conditioning embedding for MDM~\cite{MDM}, which synthesizes the corresponding motion trajectory. We apply validity checks such as root joint height bounds to remove non-physical artifacts including floating, sinking, and interpenetration, and use a vision and language model (VLM) to assess semantic alignment between prompt and motion. Segments that pass both filters are added to the synthetic training set. The pipeline repeats over multiple rounds with gradually increased difficulty, allowing the corpus to evolve toward high dynamic professional behaviors while keeping computational cost low and preserving the pretrained generator.

\subsection{Training and evaluation of Controller}
Our single primitive tracker uses reinforcement learning imitation under a single policy with dense rewards on pose, joint velocities, end effectors, and contact events, plus sample filtering, to achieve high coverage tracking and stable optimization. The same RL setup remains compatible with diverse data and feature pipelines such as PHC~\cite{luo2023perpetual} and MaskedMimic~\cite{tessler2024maskedmimic}, and we keep their original hyperparameters and training settings. The physics-based controller is trained only on our synthesized motions and is the only compute intensive component.

After convergence, we evaluate with mpjpe-g, the average joint position error in the world frame; mpjpe-l, the joint position error in the root relative frame; vel-dist, the mean per joint linear velocity discrepancy that reflects smoothness; and accel-dist, the mean per joint acceleration discrepancy that exposes high frequency jitter. These metrics delineate the controller’s capability frontier and feed back into the next iteration to guide difficulty escalation in the data generation loop.

\subsection{Competitive iteration between the controller and the dataset}
\begin{figure}
    \centering
    \includegraphics[width=0.7\linewidth]{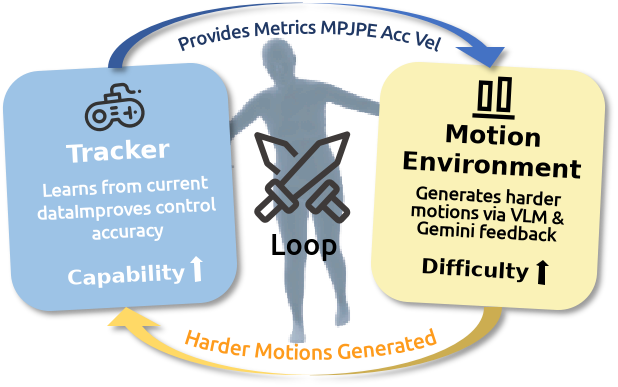}
    \caption{Competitive iteration between the controller and the dataset}
    \label{fig:compet}
\end{figure}

We employ a competitive iterative curriculum (Fig.~\ref{fig:compet}) to raise the controller performance ceiling. After each training cycle, if objective metrics exceed predefined thresholds, we treat the current distribution as mastered and escalate data difficulty. The controller then trains on harder samples, which advances its capability. Progress is driven by co-evolution of controller performance and data difficulty, and is measured by a joint evaluation that combines objective physical tracking metrics with subjective visual judgments, producing a self-reinforcing curriculum.
For closed loop feedback, we render each training motion as a concatenated SMPL sequence and evaluate it with two vision and language models, GPT‑4o and Qwen‑VL‑MAX. For each clip, the VLMs return a subjective difficulty score and descriptors covering action sequence, technical complexity, intensity, balance, continuity, plus a rationale, while the controller reports objective physical metrics on the same motion. We concatenate these signals into a semantic observation vector (obs) that captures both physical execution and visual perception.

When the observation indicates improved mastery, we raise data difficulty. Guided by our professional templates and difficulty attributes, we generate harder samples within the target domains using language prompts and MDM-based compositional synthesis. Each composition step is treated as an action that yields a new motion sequence, and the loop is automated end-to-end so data difficulty and controller competence advance together.


\subsection{Automation of the iterative process}
We instantiate a central LLM policy with a role prompt to drive a competitive closed loop curriculum where dataset difficulty co-evolves with controller performance. Concretely, we use Gemini with chain-of-thought to fuse tracker metrics, subjective visual feedback, and the previous action prompt into a semantic observation, while a VLM ensemble (GPT‑4o and Qwen‑VL‑MAX) supplies prompt–motion alignment signals. The policy outputs the next prompt from a difficulty aware variable bank grounded in our professional templates, and the environment executes motion tracking training on the synthesized clips. Optimization is implicit: the policy aims to improve physical tracking scores while steadily raising annotated difficulty, yielding a self-reinforcing curriculum. The entire loop process is illustrated in Algorithm  \ref{alg:llm_iter} and Fig.~\ref{fig:obsaction}.
\begin{figure}
    \centering
    \includegraphics[width=1\linewidth]{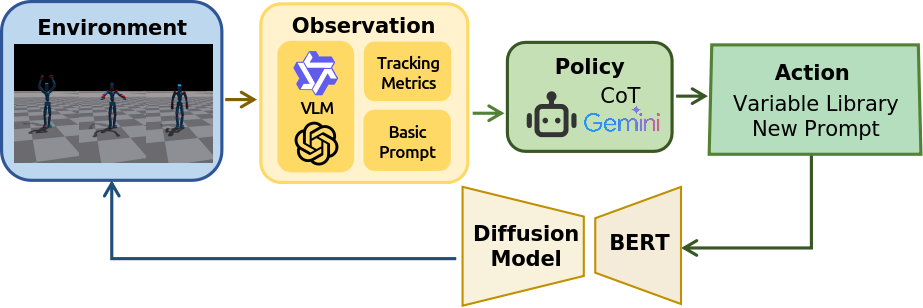}
    \caption{Schematic Diagram of the Automated Iterative Loop}
    \label{fig:obsaction}
\end{figure}
Running the loop for $K$ iterations yields a five domain corpus with $K$ difficulty tiers across martial arts, dance, combat, gymnastics, and sports, together with a physics-based controller that adapts to heterogeneous high difficulty motions. A centralized LLM with chain-of-thought uses multimodal feedback to select informative harder prompts while preserving semantic fidelity, enabling stable performance gains without pipeline specific modifications.


\begin{algorithm}[h]
\caption{\textbf{LLM-Driven Competitive Dataset--Controller Iteration}}
\label{alg:llm_iter}
\begin{algorithmic}[1]

\STATE \textbf{Input:} variable library $\mathcal{L}$, templates $\mathcal{T}$, generator $G$, VLM evaluator $F_{\text{VLM}}$, policy LLM $\pi_\theta$ (Gemini CoT).
\STATE Initialize dataset $\mathcal{D}\!\leftarrow\!\varnothing$, motion sets $\mathcal{M}\!\leftarrow\!\varnothing$, tracker $\pi^{\text{trk}}_0$, action $a_0$.

\FOR{$k=0$ \TO $K-1$}

    \STATE Compute tracking metrics $m_k$ and VLM difficulty/feedback $v_k$.
    \STATE Encode previous action $e_k\!=\!\phi(a_k)$ and form observation $o_k\!=\![m_k,v_k,e_k]$.

    \STATE Generate new action prompts:
    \[
        A_k=\{a_k^1,\dots,a_k^M\}\sim\pi_\theta(o_k,\mathcal{L},\mathcal{T}).
    \]

    \STATE Initialize $M_k\!\leftarrow\!\varnothing$.
    \FOR{$a_k^j\in A_k$}
        \STATE $q_k^j\!\leftarrow\!G(a_k^j)$; \textbf{if} fails physics \textbf{continue}.
        \STATE \textbf{if} VLM alignment is sufficient: $M_k\!\leftarrow\!M_k\cup\{(q_k^j,a_k^j)\}$.
    \ENDFOR

    \STATE $\mathcal{D}\!\leftarrow\!\mathcal{D}\cup M_k$;\quad $\pi^{\text{trk}}_{k+1}\!\leftarrow\!\text{TrainTracker}(\mathcal{D})$.
    \STATE Store motion set: $\mathcal{M}\!\leftarrow\!\mathcal{M}\cup\{M_k\}$.
    \STATE Update summary prompt $a_{k+1}$.

\ENDFOR

\STATE \textbf{Return:} best tracker $\pi^{\text{trk}}_*$ and motion sets $\mathcal{M}$.

\end{algorithmic}
\end{algorithm}

\section{Experiments}
\label{sec:experiments}
Our experiment aims to address the following core question:
(1) Do our definitions and generation extrapolate beyond MDM’s training sets (AMASS, HumanAct12)?
(2) Does the framework achieve generalization to unseen high-dynamics and high-difficulty data?
(3) Does the feedback loop raise difficulty over time and improve controller stability and performance?

\subsection{Experimental Setup}

All experiments are performed on a single NVIDIA A6000 GPU. We use the PHC single-primitive training configuration~\cite{luo2023perpetual} as the base setting for reinforcement learning. Each loop iteration trains a physics-based tracker on the synthetic motion dataset generated from the current prompt batch. Unless otherwise noted, we report metrics on six standard motion test sets: kungfu, emdb, amass, mdm, aist++, and video-converted.

\subsection{Motion professionalism verification}
\begin{figure}[t]
  \centering
  \begin{subfigure}[t]{0.23\textwidth}
    \centering
    \includegraphics[width=\linewidth]{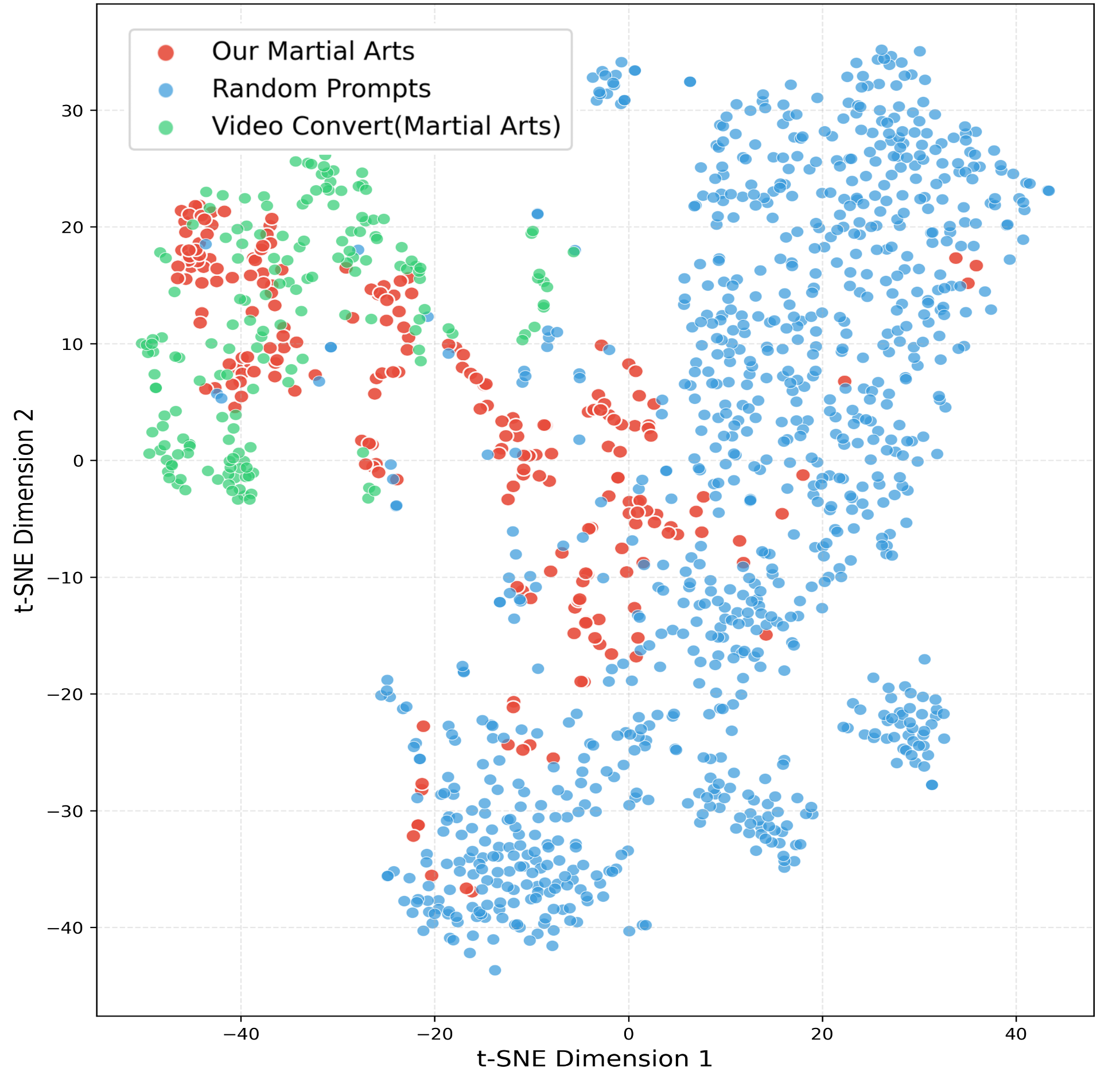}
    \caption{}
    \label{fig:tsne-a}
  \end{subfigure}\hfill
  \begin{subfigure}[t]{0.23\textwidth}
    \centering
    \includegraphics[width=\linewidth]{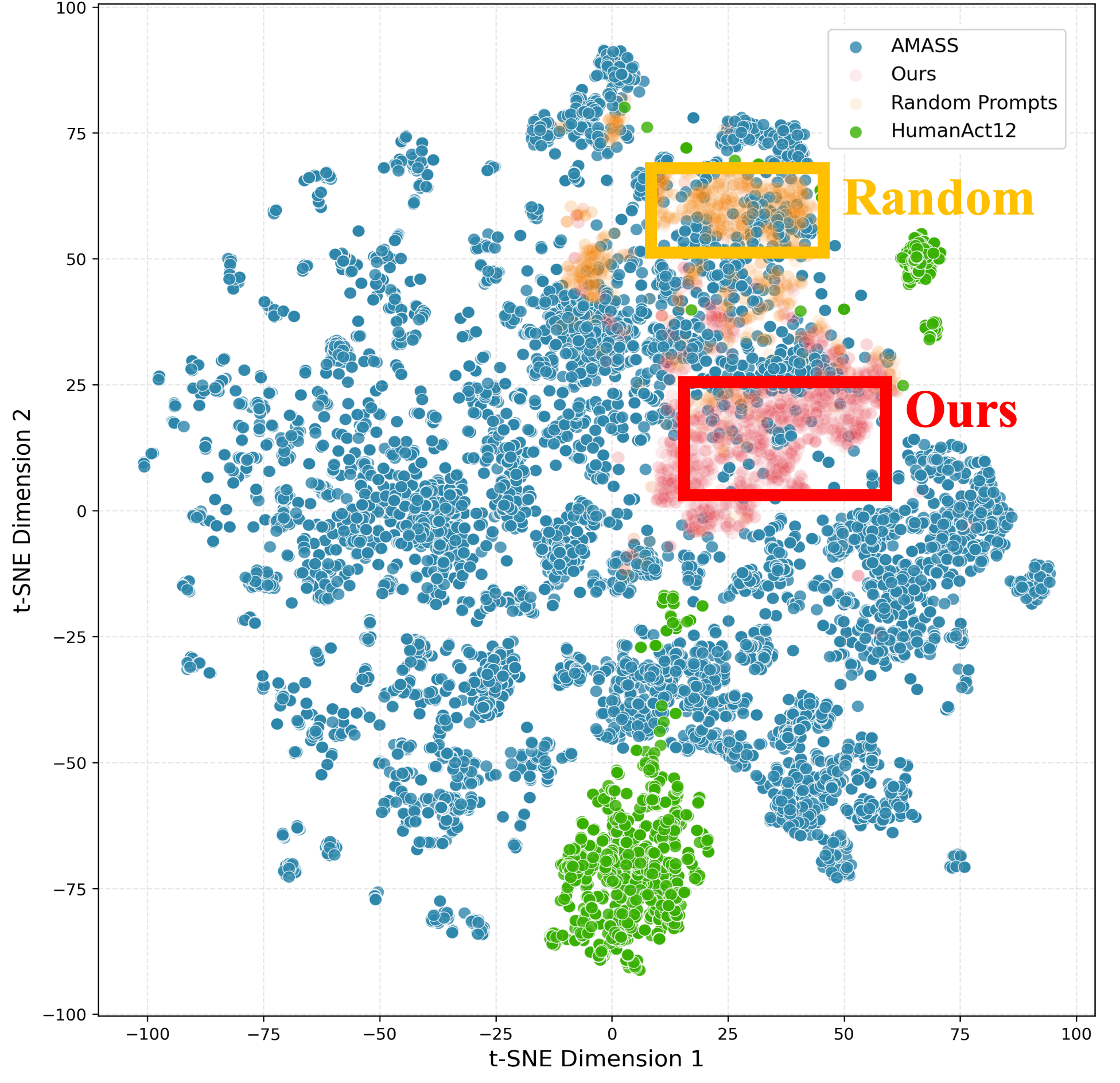}
    \caption{}
    \label{fig:tsne-b}
  \end{subfigure}
  \caption{(a) The t-SNE graph of martial arts datasets comparison; (b) The t-SNE graph of all datasets(AMASS/Ours/Random Prompts/HumanAct).}
  \label{fig:obsaction}
\end{figure}

\begin{figure}
    \centering
    \includegraphics[width=0.9\linewidth]{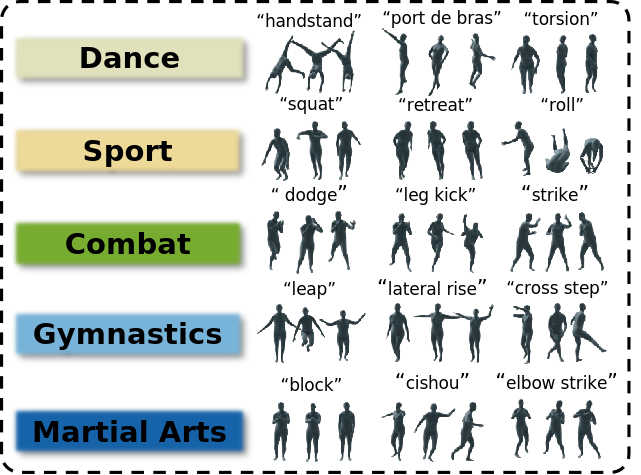}
    \caption{Visualization of 5 categories of professional movements}
    \label{fig:5kinds}
\end{figure}

\paragraph{Motion dataset features.}We build training data by combinatorially instantiating a variable prompt library, synthesizing motions with MDM-step50-BERT, rendering poses via SMPL, and filtering with a VLM; Fig.~\ref{fig:tsne-a} illustrates representative prompt categories and shows semantic alignment between action terms and rendered clips(Fig.~\ref{fig:5kinds}). To assess domain specificity, we compare (i) professionally curated martial-arts motions, (ii) motions synthesized from our expert martial arts prompts, and (iii) motions from Gemini-generated random prompts. Using per-frame mean global joint rotation features and t-SNE visualization, expert prompt syntheses substantially overlap the professional martial-arts manifold, while random-prompt motions lie far away. This indicates our prompt design encodes salient domain priors and mitigates distributional bias in MDM prompt-conditioned synthesis.


\begin{figure}[t]
    \centering
    \includegraphics[width=0.8\linewidth]{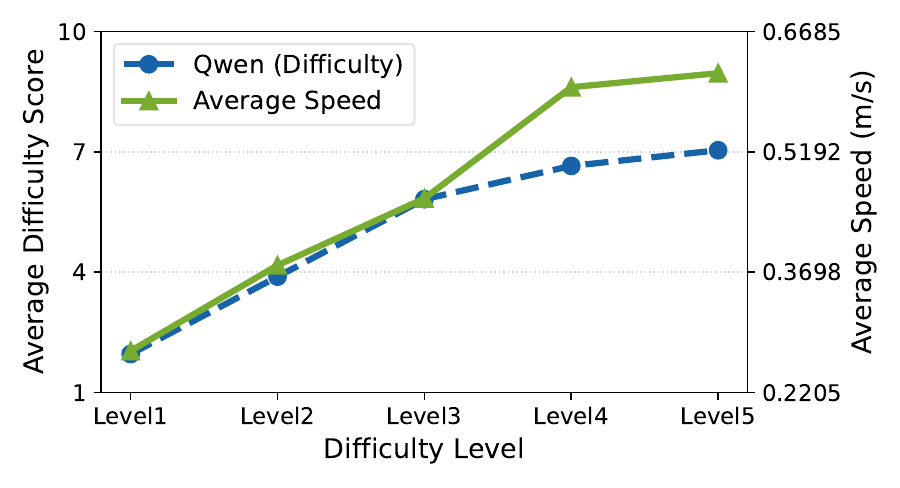}
    \caption{Loop-wise Difficulty and Speed Trends of Qwen Evaluations} 
    \label{fig:qwenbelieve} 
\end{figure}

\begin{table}[t]
  \centering
  \begin{threeparttable}
    \caption{Success rate (\%) of different pipelines across test sets (totaling 2201 clips). The test suite includes: Motion-X/Kungfu (663), EMDB (45), AIST++ (1320), and Video-Convert (173). We report per-set results and the average (Avg) calculated by clip.}
    \label{tab:succ vs baseline}
    \small
    \setlength{\tabcolsep}{5pt}
    \begin{tabular}{lccccc}
      \toprule
      Method & Kungfu & EMDB & AIST++ & VC & Avg \\
      \midrule
      AMASS  & 47.1 & 53.3 & 67.6 & 31.2 & 58.3 \\
      L0     & 37.8 & 31.1 & 68.8 & 33.3 & 55.9 \\
      L1     & 47.7 & 33.3 & 75.3 & 38.7 & 64.0 \\
      L2     & 51.7 & 51.1 & 73.3 & 41.6 & 63.8 \\
      L3     & 59.1 & 64.4 & 82.1 & 50.9 & 72.4 \\
      L4     & 55.8 & 55.6 & 84.0 & 45.1 & 71.8 \\
      L5     & \textbf{60.6} & 60.0 & 85.2 & 54.3 & 74.8 \\
      L6     & 60.3 & \textbf{64.4} & \textbf{88.1} & \textbf{58.9} & \textbf{76.9}$^{\dagger}$ \\
      \bottomrule
    \end{tabular}
    \begin{tablenotes}
      \footnotesize
      \item[$^{\dagger}$] Compared to the AMASS baseline, our L6 model reduces the average failure rate by 45\% (from 41.7\% to 23.1\%).
    \end{tablenotes}
  \end{threeparttable}
\end{table}
\paragraph{Motion dataset and MDM relationship.}
We run a t-SNE analysis on clip-level pose embeddings to test whether our prompt-conditioned synthesis departs from MDM’s training distribution. The corpus includes HumanML3D training data (AMASS, HumanAct12), motions synthesized by MDM from Gemini-generated random prompts, and motions synthesized by MDM from our expert prompts. Fig.~\ref{fig:tsne-b} shows our expert-prompt samples lie largely outside the MDM training manifold, while random-prompt samples cluster within it—indicating MDM can produce out-of-distribution motions and that expert prompts effectively elicit novel content beyond the original dataset coverage.

\subsection{Generalization research}
\begin{figure*}
    \centering
    \includegraphics[width=1\linewidth]{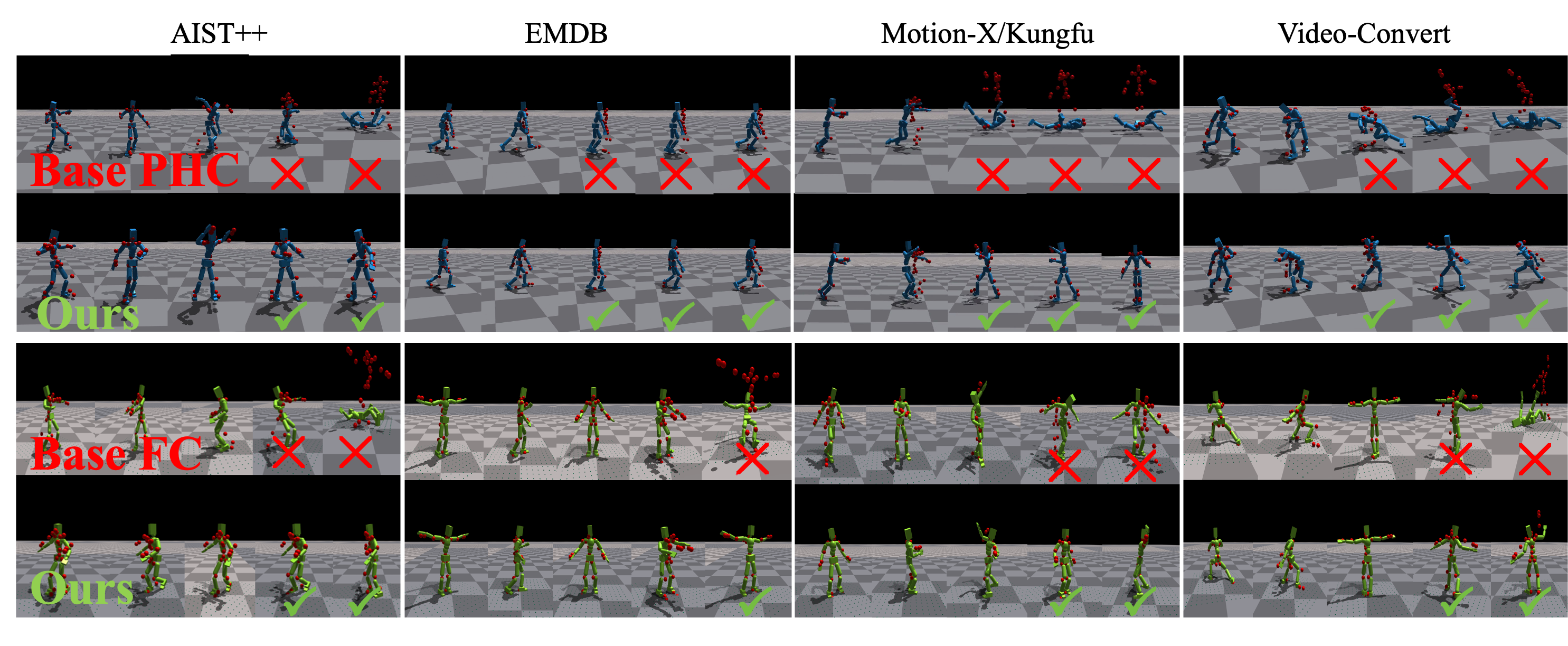}
    \caption{Qualitative tracking performance between Ours and Baselines of PHC and Maskedmimic.}
    \label{fig:controller}
\end{figure*}

\paragraph{OOD dataset testing.}
We evaluate portability by instantiating our iterative framework with the PHC single-primitive controller trained on AMASS and comparing to a non-iterative baseline under matched protocols. The iterative pipeline synthesizes motions with MDM-step50-BERT, applies root-height and VLM-based filtering to form loop0, trains PHC, and collects tracking metrics. Using SMPL renderings and VLMs for difficulty/attribute signals, we construct an observation passed to a Gemini CoT module to propose harder, controller-targeted prompts; repeating synthesis and filtering yields loop1 and subsequent loops. On four high-difficulty third-party benchmarks (Tab.~\ref{tab:succ vs baseline}), loop0 lags the baseline, loop1 already surpasses it, and later loops continue improving—showing the feedback-driven, prompt-conditioned curriculum is plug-and-play with the PHC single-primitive paradigm and outperforms one-shot training on large, generic corpora.
\paragraph{Compatibility of different trackers.}
We test cross-paradigm effectiveness by instantiating our iterative framework with the DeepMimic-based MaskedMimic controller, comparing to a baseline MaskedMimic (FC) trained on AMASS in IsaacGym. Our method uses loop0 set, trains the controller, collects PHC-style inference-time metrics, and uses Gemini to generate harder loop1 prompts for data synthesis and continued training. On third-party, high-difficulty suites (Tab.~\ref{tab:maskedmimic_sr} and Fig.\ref{fig:controller}), loop0 matches the baseline, while loop1 delivers consistent, significant improvements—showing that feedback-driven, prompt-conditioned data-and-policy refinement transfers to distinct tracking-by-imitation frameworks.

\begin{table}[htbp]
\centering
\small
\caption{Maskedmimic FC*: Success Rate on each test set."*" means early period.}
\begin{tabular}{lccc}  
\toprule
\textbf{Dataset} & \textbf{AMASS} & \textbf{loop0} & \textbf{loop1} \\  
\midrule
Motion-X/Kungfu  & 57.2           & 54.0           & \textbf{65.8}           \\  
EMDB             & 53.3           & 48.9           & \textbf{71.1}           \\
AIST++           & 68.9           & 75.3           & \textbf{83.9}           \\
Video-Convert    & 41.6           & 47.4           & \textbf{62.4}           \\  
\bottomrule
\end{tabular}
\label{tab:maskedmimic_sr}
\end{table}
\subsection{Ablation Study}
\begin{figure}
    \centering
    \includegraphics[width=0.8\linewidth]{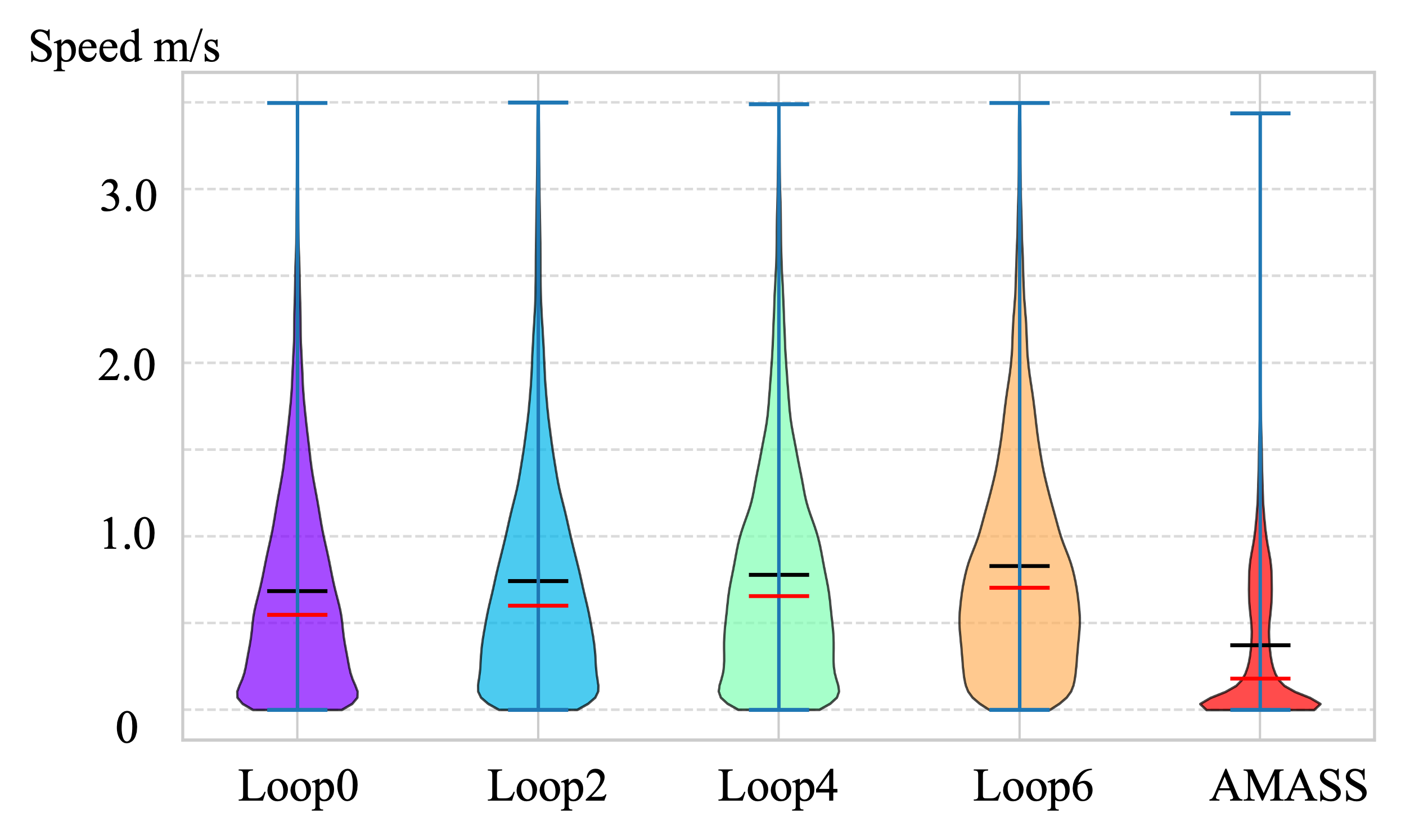}
    \caption{The velocity distribution of AMASS and ours.}
    \label{fig:vio}
\end{figure}

\paragraph{VLM scoring reliability verification.}
To ensure an objective verification, we construct five tiers of compositional test prompts from predefined variable libraries (Levels 1–5) and instantiate 200 motions per tier (N=1,000). For each motion, we compute the mean velocity as a coarse physical difficulty proxy and render frame sequences. A vision–language model, Qwen, blinded to the text prompts and given only the rendered frames, assigns a difficulty score on a 1–10 scale. As shown in Fig.~\ref{fig:qwenbelieve}, both the prompt-defined difficulty level and the physical metric exhibit a clear monotonic trend: average velocity increases with tier, and Qwen’s difficulty ratings rise accordingly. This alignment across prompt design, physics-based indicators, and VLM-based assessments supports the credibility of the automated VLM scoring.
\paragraph{Observation role ablation.}
We ablate the observation vector in our iterative feedback pipeline on the PHC single-primitive controller, fixing the Gemini-CoT difficulty objective and comparing: (1) full (physics metrics + VLM), (2) w/o VLM, (3) w/o physics metrics, and (4) w/o both. Starting from the same loop0, each variant is trained through `loop3` and evaluated by average success on four third-party, high-difficulty benchmarks (Tab.~\ref{tab:ablation_var}). Results are consistent: no observations $<$ no physics $<$ no VLM $<$ full. This indicates the scheduler becomes stochastic without observations; physics-based tracking metrics are the most diagnostic for targeted challenge design, while VLM feedback provides complementary, subjective difficulty cues. Both signals help, with physics being critical and VLM offering additional gains.
\begin{table*}[t]
  \centering
  \caption{Ablation study on success rate across different test sets and loops.}
  \label{tab:ablation_var}
  \small
  \begin{tabular}{l|ccc|ccc|ccc|ccc|ccc}
    \toprule
    \multirow{2}{*}{Method} &
      \multicolumn{3}{c|}{Motion-X/Kungfu} &
      \multicolumn{3}{c|}{EMDB} &
      \multicolumn{3}{c|}{AIST++} &
      \multicolumn{3}{c|}{Video-Convert}\\
    & L1 & L2 & L3
    & L1 & L2 & L3
    & L1 & L2 & L3
    & L1 & L2 & L3\\
    \midrule
    full (var) &
      47.7 & 51.7 & 59.1 &
      33.3 & 51.1 & \textbf{64.4} &
      75.3 & 73.3 & \textbf{82.1} &
      38.7 & 41.6 & 50.9 \\
    w/o var &
      53.5 & 59.1 & \textbf{59.9} &
      53.3 & 57.8 & 53.3 &
      75.6 & 80.2 & 76.5 &
      50.3 & 48.0 & 46.8 \\
    w/o VLMs &
      46.3 & 54.8 & 58.4 &
      42.2 & 62.2 & 60.0 &
      73.8 & 80.0 & 81.9 &
      43.4 & 45.1 & \textbf{57.2} \\
    w/o metrics &
      48.0 & 50.7 & 54.6 &
      31.1 & 31.1 & 35.6 &
      78.3 & 77.5 & 78.9 &
      38.2 & 28.9 & 40.5 \\
    w/o VLMs+metrics &
      46.5 & 54.3 & 54.3 &
      35.6 & 46.7 & 46.7 &
      71.0 & 76.1 & 77.4 &
      38.2 & 46.8 & 50.9 \\
    \bottomrule
  \end{tabular}
\end{table*}
\paragraph{Variable template library role verification.}
To assess the contribution of the variable library, we conduct an ablation in which it is removed and Gemini-CoT is tasked with proposing harder, expert-oriented prompts directly. The `loop0` model and training protocol remain unchanged; we then iterate to `loop3`. As summarized in Tab.~\ref{tab:ablation_var}, this variant underperforms the configuration with the variable library across all third-party benchmarks. These results indicate that the variable library provides a structured prior that stabilizes prompt generation, yielding more targeted and diagnostically challenging motions. By improving the specificity and controllability of the synthesized data, the library enables Gemini-CoT to more reliably escalate task difficulty and thus push the controller’s performance frontier.
\begin{table}[htbp]
\centering
\small %
\caption{Metrics(Success Rate) on each dataset for 1400 clips testset Loop0 and our Loop6.}
\begin{tabular}{l|c|c}
\toprule
\multirow{2}{*}{\textbf{Dataset}} &
\multicolumn{1}{c|}{\textbf{Loop0(1400 clips)}} &
\multicolumn{1}{c}{\textbf{Loop6}} \\
 & \textbf{SR} & \textbf{SR} \\
\midrule
Kungfu(663)     & 58.7 & \textbf{60.3} \\
EMDB(45)         & \textbf{73.3} & 64.4 \\
AIST++(1320)        & 85.3 & \textbf{88.1} \\
Video-Convert(173) & 55.5 & \textbf{59.0} \\
\bottomrule
\end{tabular}
\label{tab:loop0_loop6_metrics}
\end{table}
\paragraph{Feedback iteration role verification.}
To isolate the effect of feedback-driven iteration from mere data scale, we construct a size-matched, non-iterative baseline. Concretely, we randomly sample from the variable library a training set whose cardinality equals the cumulative data used across loop0 to loop6, and train a single model to convergence under the same optimization protocol. As reported in Tab.~\ref{tab:loop0_loop6_metrics}, despite identical data budgets, our feedback-iterative regimen (`loop0`→`loop6`) consistently outperforms the one-shot baseline across multiple large-scale test suites. This evidences that the gains stem from iterative feedback and curriculum effects rather than simply from increasing the number of training motions.
\paragraph{Dataset difficulty variation verification}
Using the pretrained composition controller PHC+, we run inference on each loop’s motion set and observe a monotonic drop in tracking success with increasing loop index (Tab.~\ref{tab:loop_metrics}), indicating a feedback-driven curriculum that yields systematically harder motions. Velocity analyses—global root-velocity distributions (Fig.~\ref{fig:vio}) and per-clip mean inter-frame speed on length-matched 180-frame sequences—show that AMASS is predominantly low-dynamics, whereas our synthesized data exhibits progressively higher velocities, broader tails, higher-frequency variations, and higher peak speeds. These trends confirm that later-loop datasets are intrinsically higher-dynamic and higher-difficulty, aligning with the observed PHC+ performance drop.

\subsection{Limitations}
CLAIMS faces two constraints. First, synthesis quality is bounded by the generative model's capacity on extreme high-dynamics; however, our modular architecture enables seamless integration of future generators. Second, the manually curated variable library lacks objective calibration and comprehensive coverage, necessitating future work on automated, multi-modal libraries with domain knowledge graphs to scale capabilities.
\begin{table}[t]
  \centering
  \caption{PHC+ tracking performance on our datasets.A lower success rate indicates that the motions are more challenging for third-party trackers.}
  \label{tab:loop_metrics}
  \small
  \begin{tabular}{lcccc}
    \toprule
    Data Loop &
    SR  &
    g-MPJPE  &
    Acc  &
    Vel  \\

    \midrule
    L0 & 75.3& 49.78 & 5.97 &  8.54  \\
    L1 & 65.8 & 53.84 & 6.66 &  9.34  \\
    L2 & 65.2& 61.29 & 7.65 & 10.86  \\
    L3 & 61.2 & 57.24 & 7.03 &  9.95  \\
    L4 & 59.0  & 57.70 & 7.08 &  9.99 \\
    L5 & 52.7 & 59.10 & 7.49 & 10.65  \\
    L6 & 53.6 & 59.61 & 7.94 & 10.97  \\
    \bottomrule
  \end{tabular}
\end{table}
\section{Conclusion}
This work targets the core bottlenecks in training data and generalization for physics‑based humanoid control. We propose a low‑cost, adaptive, closed‑loop pipeline for motion data generation and iterative controller optimization, offering a new path to high‑dynamics professional motion control. The proposed framework cuts data generation cost, boosts controller generalization, and establishes a data–controller co‑evolution paradigm for advancing physics‑based humanoid control toward high dynamics and diverse scenarios.

\section*{Acknowledgements}
This work was supported by the Red Bird MPhil Program at The Hong Kong University of Science and Technology (Guangzhou).
{
    \small
    \bibliographystyle{cvpr}
    \bibliography{ref}
}



\end{document}